# Neighborhood Rough Set based Multi-document Summarization

Nidhika Yadav*

*Abstract:* This research paper proposes a novel Neighbourhood Rough Set based approach for supervised Multi-document Text Summarization (MDTS) with analysis and impact on the summarization results for MDTS. Here, Rough Set based LERS algorithm is improved using Neighborhood Rough Set which is itself a novel combination called Neighborhood-LERS to be experimented for evaluations of efficacy and efficiency. In this paper, we shall apply and evaluate the proposed Neighborhood-LERS for Multi-document Summarization which here is proved experimentally to be superior to the base LERS technique for MDTS.

## 1. Introduction

Learning from examples using Rough Sets has been a popular technique both theoretically and for experimental works too. The theoretical work in this research area of Rough Sets is prevalent primarily for development of new concepts which can improve on the results of specific framework of certain problems. While, the experimental analysis has its own parallel importance in order to establish, the use of a particular technique on some preferred datasets by that technique. Further, experimental analysis has gained even more emphasis than theoretical efforts, in this specific area, in order to get the learned parameters for one particular problem at hand.

In this research work, we focus on the problem of Multi Document Text Summarization (MDTS) [2, 3, 6]. This topic is very important from point of view of research and results evaluations as well. The evolution of MDTS started when the processing units which can be a super computer or personal computers, or any other computing devises, were capable of handling single documents processing. Since typically any kind of work requires multiple text files, be it an organizations data or be it an online content with hyperlinks to other html files or be it a file of texts in a legal database, or a collection of research articles to be processed via a computer. All these examples show that there was a great need to handle multiple text files, and not just a single text file. Though single document summarization has its own uses. Multiple document summarization mainly evolved due to plethora of text documents appearing on web, data repositories, records, newspapers to mention a few. Further text data itself may be present in many formats such as text embedded in html files, pdf files, word documents, open office format or simply a text file.

Rough Sets [14, 15] on the other hand provides a good source of techniques which can be used in various phases of problem solving. They typical problems which require the use of Rough Sets has been the cases where uncertainty has to be dealt with. Rough Set based algorithms deals with uncertain data and uncertainty in problem solving too. It has been used in a wide variety of applications but still till recent times there was no work in area of Text Summarization. Lately, Yadav and Chatterjee have proposed efficient techniques for the same for Single [5, 13, 14] and Multi Document Summarization [15]. The key motivation of this paper is to improve the results obtained by the research works which deal with uncertainty and MDTS. The novel technique developed here was motivated by use of LERS algorithm of Rough Sets to compute supervised summarization of multiple documents. After analysis of LERS algorithm, the key drawback was a need of nominal attributes, while the attributes computed are primarily numeric. Hence, a hybrid technique viz. neighbourhood Rough Set based LERS algorithm was proposed and experiments were performed for its use in MDTS. We have taken as baselines the work by Yadav and Chatterjee [15] and have attained much improvements on the work. We shall be focused with supervised MDTS [4] in this work.

---

*Nidhika Yadav have recently completed PhD from IIT Delhi. This work was performed in IIT Delhi during her Ph.D.

## 2. The Proposed Neighborhood LER Algorithm

The key drawback of the LERS algorithm is that it requires categorical data to compute the rules. While, major problems occurring in nature contain numerical values generated though algorithms, collected through sensors or obtained through sources such as measurements. Also, it may be noted that the Information System used in this work is numeric valued. We propose to modify the existing LERS algorithm using Neighborhood Rough Sets [16]. Firstly, we define Neighborhood Rough Sets as follows.

Neighborhood Rough Set theory (Yao, 1999) [16], is based on the indiscernibility relation created using a distance metric to estimate the similarity of two objects for a Neighborhood radius. Consider the Information System $(U, A)$ where $U$ is the Universe under consideration and $A$ is the knowledge about $U$ in the form of the set of attributes $\{a1, a2, ... an\}$. For an element $x \in U$ the $\delta$-neighborhood of $x$ is given as [16]:

$$\delta_P(x) = \{ y \mid y \in U, \Delta^P(x,y) \leq \delta \}$$

where, $\Delta^P$ is the distance function. A generic metric for distance functions is Minkowsky distance:

$$\Delta^P(x,y) = \left( \sum_{i=1}^{n} | f(x, ai) - f(y, ai)|^P \right)^{1/P}$$

Where $f(x, a)$ is value of $x$ for attribute value $a$. The following are specialized and popular forms of Minkowsky distance metric.

    i.    $\Delta^P$ is Manhattan distance if $P = 1$.

    ii.    $\Delta^P$ is Euclidean distance if $P=2$.

    iii.    $\Delta^P$ is called Chebychev distance if $P = \infty$

The upper and lower approximations defined in terms of Neighborhoods are given as follows (Hu et al, 2008):

$$\underline{P}X = \{ x \in U : \delta_P(x) \subseteq X\}$$
$$\overline{P}X = \{ x \in U : \delta_P(x) \cap X \neq \emptyset\}$$

The other definitions such as boundary region and negative region can be defined as before using these new definitions of lower and upper approximations.

We here propose Neighborhood Rough Set based Learning from examples which shall be referred as NLER algorithm. The algorithm is given in Fig. 1 and is based on LER algorithm.

---
**ALGORITHM.**   Proposed NLER Algorithm for Text Summarization

---
**Input:** Training and Testing Data in form of document collection and $N$ = number of words required in summary.
**Output:** Extractive summary
**Start**
1. Compute target class $D$ for training data using ROUGE scores.
2. Parse the sentences in training data using and compute the features $P$ of the sentences.
3. Add the sentences and all features to Information System $(U, P)$.
4. Determine the Global Covering call it $A$ using Neighborhood Rough Sets based distance for each feature viz. each feature should be in a $\alpha$-neighborhood for a specific $\alpha$.
5. **If** $A^* \leq D^*$
   **then**
      5.1  Generate the Global Covering and rules.
      5.2  Reduce the rules further by deleting attributes from rules.
      5.3  Check for consistency of reduced rules.
   **End**

6. Output the rules
   7. Test the rules on testing data and generate importance of sentences.
   8. Create the *N* words summary by selecting *N* words form beginning of text.
   **End**

**Fig. 1. Algorithm for NLER based MDTS**

Here the lower approximation and upper approximation are computed using Neighborhood Rough Set. The attributes which correspond to the global coverings are selected. Further, each attribute-value pair is dropped from a rule and resulting rule set is checked for consistency using Neighborhood Rough Set. The minimal number of attribute-value pairs are kept in the final rule. Finally, this rule set is used to evaluate the testing phase.

In Section 3 the proposed Rough Set based techniques are evaluated with the popular supervised techniques and the experimental results are analysed.

3. **Concluding Experiments & Results of Proposed Techniques**

In this section we analyse the techniques namely (i) Neighbourhood Rough Sets based supervised learning, (ii) Rough Set based LERS, (iii) Fuzzy Rough Set based Nearest Neighbour Algorithm and (iv) other non-Rough Set based supervised techniques. Moreover, these techniques are evaluated with and without post-processing with Aggregate-Rank-Measure [15] and with classical membership [15] based post-processing to get the conclusions of the results of experiments conducted.

The non-Rough Set based supervised techniques that are evaluated for performance are: Fuzzy Nearest Neighbour, K-Nearest Neighbour, Logistic Regression, Naïve Bayes, Neural Networks, Random Forest and Support Vector Machine [4]. The proposed Neighborhood Rough Set based LERS algorithm is evaluated with a Neighborhood radius of 0.2 and 0.3 and these are represented as Nbhd_0.2 and Nbhd_0.3 respectively.

The datasets used for training consists of documents from DUC2002 corpus [7]. Each cluster consists of five to ten documents. DUC2003 [7] and DUC2005 datasets [7] have been used for evaluation of the results. From DUC2003 datasets sixty document clusters each made up of multiple Single documents have been used for evaluations. We considered DUC2005 dataset consisting of 50 large document clusters of news articles for evaluation. The results were computed by comparing the extracts with the gold summaries created by experts.

|  | FuzzyNN [15] | Fuzzy Rough NN [15] | KNN [15] | Logistic [15] | Naïve Bayes [15] | Nbhd 0.2 | Nbhd 0.3 | Neural [15] | Random Forest [15] | SVM [15] | LERS [15] |
|---|---|---|---|---|---|---|---|---|---|---|---|
| **ROUGE-1** | 0.2798 | 0.2815 | 0.2794 | 0.2599 | 0.2772 | 0.2849 | 0.2812 | 0.2726 | 0.2476 | 0.2654 | **0.2952** |
| **ROUGE-2** | 0.0482 | 0.0480 | 0.0478 | 0.0447 | 0.0457 | 0.0489 | 0.0475 | 0.0441 | 0.0412 | 0.0460 | **0.0548** |
| **ROUGE-L** | 0.2349 | 0.2369 | 0.2354 | 0.2191 | 0.2338 | 0.2394 | 0.2375 | 0.2330 | 0.2125 | 0.2240 | **0.2459** |
| **ROUGE-SU** | 0.0866 | 0.0863 | 0.0853 | 0.0798 | 0.0852 | 0.0877 | 0.0868 | 0.0829 | 0.0761 | 0.0815 | **0.0926** |

**Table 1. Results of DUC2003 for various supervised learning algorithms and Neighborhood Rough Sets without Aggregate-Rank-Measure**

|  | FuzzyNN [15] | Fuzzy Rough NN [15] | KNN [15] | Logistic [15] | Naïve Bayes [15] | Nbhd 0.2 | Nbhd 0.3 | Neural [15] | Random Forest [15] | SVM [15] | LERS [15] |
|---|---|---|---|---|---|---|---|---|---|---|---|
| ROUGE-1 | 0.2640 | 0.2716 | 0.2666 | 0.2681 | 0.2667 | 0.2628 | 0.2628 | 0.2522 | 0.2629 | 0.2698 | 0.2640 |
| ROUGE-2 | 0.0425 | 0.0453 | 0.0427 | 0.0436 | 0.0422 | 0.0426 | 0.0426 | 0.0370 | 0.0390 | 0.0444 | 0.0425 |
| ROUGE-L | 0.2640 | 0.2716 | 0.2666 | 0.2681 | 0.2667 | 0.2628 | 0.2628 | 0.2522 | 0.2629 | 0.2698 | 0.2640 |
| ROUGE-SU | 0.0866 | 0.0863 | 0.0853 | 0.0798 | 0.0852 | 0.0877 | 0.0868 | 0.0829 | 0.0761 | 0.0815 | 0.0926 |

**Table 2. Results of DUC2003 for various supervised learning algorithms and Neighborhood Rough Sets with classical membership functions**

The results for all the supervised techniques and Neighborhood Rough Sets models for summarization of DUC2003 documents without post-processing with Aggregate-Rank-Measure are provided in Table 1. The results show that best results are obtained for LERS algorithm for all measures of ROUGE scores.

|  | FuzzyNN [15] | Fuzzy Rough NN [15] | KNN [15] | Naïve Bayes [15] | Nbhd 0.2 | Nbhd 0.3 | Neural Network [15] | Random Forest [15] | SVM [15] | LERS [15] |
|---|---|---|---|---|---|---|---|---|---|---|
| ROUGE-1 | 0.2859 | 0.2863 | 0.2832 | 0.2884 | **0.2915** | 0.2910 | 0.2689 | 0.2623 | 0.2770 | 0.2796 |
| ROUGE-2 | 0.0375 | 0.0363 | 0.0353 | 0.0368 | **0.0391** | 0.0386 | 0.0333 | 0.0337 | 0.0348 | 0.0375 |
| ROUGE-L | 0.2386 | 0.2377 | 0.2344 | 0.2382 | **0.2396** | 0.2396 | 0.2234 | 0.2189 | 0.2304 | 0.2336 |
| ROUGE-SU | 0.0870 | 0.0865 | 0.0854 | 0.0871 | **0.0884** | 0.0884 | 0.0812 | 0.0796 | 0.0668 | 0.0861 |

**Table 3. Results of DUC2005 for various supervised learning algorithms and Neighborhood Rough Sets without Aggregate-Rank-Measure**

Table 2 gives results for DUC2003 datasets with post-processing with classical membership function. It is evident from the results given here that no improvement in results are noted when instead of post-processing with Aggregate-Rough-Sets we post-process with classical Rough Set based membership. Table 3 provide results with Aggregate-Rank-Measure applied on the top of all the supervised summarization techniques, both Rough Set based and non-Rough Set based, for DUC2003 dataset. It is evident that Rough Set based techniques performs better for DUC2003 dataset both with and without Aggregate-Rank-Measure. The proposed Neighborhood Rough Sets model (NLER), namely Nbhd_0.2, with neighborhood radius of 0.2 is found to be performing the best for ROUGE-1, ROUGE-L and ROUGE-SU metrics while LERS performs best for ROUGE-2 scores. There is increase in ROUGE scores on using Aggregate-Rank-Measure in DUC2003 documents and this increase in scores is persistent for all the supervised techniques analysed. Tables 4 presents the results for DUC2005 dataset with Neighborhood Rough Sets models. No post-processing has been performed here. It is observed that proposed Neighborhood Rough Sets models (NLER) namely Nbhd_0.2 and Nbhd_0.3 with neighborhood radius of 0.2 and 0.3 have been found to be performing the best for ROUGE-1, ROUGE-2, ROUGE-L and ROUGE-SU metrics.

|  | FuzzyNN [15] | Fuzzy Rough NN | KNN [15] | Logistic [15] | Naïve Bayes [15] | Nbhd 0.2 | Nbhd 0.3 | Neural [15] | Random Forest [15] | SVM [15] | LERS [15] |
|---|---|---|---|---|---|---|---|---|---|---|---|
| ROUGE-1 | 0.2940 | 0.2956 | 0.2954 | 0.2750 | 0.2858 | **0.2985** | 0.2955 | 0.2903 | 0.2647 | 0.2789 | 0.2952 |
| ROUGE-2 | 0.0508 | 0.0517 | 0.0518 | 0.0483 | 0.0476 | 0.0522 | 0.0486 | 0.0493 | 0.0460 | 0.0490 | **0.0548** |
| ROUGE-L | 0.2448 | 0.2448 | 0.2448 | 0.2448 | 0.2448 | 0.2448 | 0.2448 | 0.2448 | 0.2448 | 0.2448 | **0.2448** |
| ROUGE-SU | 0.0920 | 0.0919 | 0.0921 | 0.0856 | 0.0877 | **0.0928** | 0.0899 | 0.0902 | 0.0829 | 0.0866 | 0.0926 |

**Table 4. Results of DUC2003 for various supervised learning algorithms and Neighborhood Rough Sets with Aggregate-Rank-Measure**

|  | FuzzyNN [15] | Fuzzy Rough NN [15] | KNN [15] | Naïve Bayes [15] | Nbhd 0.2 | Nbhd 0.3 | Neural Network [15] | Random Forest [15] | SVM [15] | LERS [15] |
|---|---|---|---|---|---|---|---|---|---|---|
| **ROUGE-1** | 0.2627 | 0.2738 | 0.2656 | 0.2682 | 0.2697 | 0.2710 | 0.2710 | 0.2656 | 0.2657 | 0.2711 |
| **ROUGE-2** | 0.0275 | 0.0299 | 0.0287 | 0.0302 | 0.0296 | 0.0324 | 0.0324 | 0.0297 | 0.0295 | 0.0308 |
| **ROUGE-L** | 0.2163 | 0.2259 | 0.2183 | 0.2207 | 0.2224 | 0.2271 | 0.2271 | 0.2184 | 0.2203 | 0.2234 |
| **ROUGE-SU** | 0.0870 | 0.0800 | 0.0773 | 0.0780 | 0.0787 | 0.0815 | 0.0815 | 0.0771 | 0.0772 | 0.0798 |

**Table 5. Results of DUC2005 for various supervised learning algorithms and Neighborhood Rough Sets with classical membership**

Table 5 gives results for post-processing with classical Rough Set based membership function for DUC2005 dataset. It is clear that the post-processing with the classical membership has not improved the results. Moreover, the results in Table 6 for DUC2005 dataset establish that there is a significantly good increase in performance after post-processing with the proposed Aggregate-Rank-Measure. Further, the proposed NLER has been found to have performed better for DUC2005 dataset than other models evaluated after the proposed post-processing. Rough Set based techniques are found to be performing best with or without post-processing among all the supervised techniques evaluated for DUC2005.

|  | FuzzyNN [15] | Fuzzy Rough NN [15] | KNN [15] | Naïve Bayes [15] | Nbhd 0.2 | Nbhd 0.3 | Neural Network [15] | Random Forest [15] | SVM [15] | LERS [15] |
|---|---|---|---|---|---|---|---|---|---|---|
| **ROUGE-1** | 0.3064 | 0.3096 | 0.3054 | 0.2893 | 0.3058 | **0.3164** | 0.3104 | 0.2692 | 0.2811 | 0.2900 |
| **ROUGE-2** | 0.0418 | 0.0404 | 0.0418 | 0.0392 | 0.0429 | **0.0432** | 0.0409 | 0.0334 | 0.0392 | 0.0378 |
| **ROUGE-L** | 0.2513 | 0.2506 | 0.2493 | 0.2367 | 0.2494 | **0.2590** | 0.2529 | 0.2239 | 0.2319 | 0.2366 |
| **ROUGE-SU** | 0.0941 | 0.0939 | 0.0933 | 0.0883 | 0.0944 | **0.0983** | 0.0947 | 0.0814 | 0.0873 | 0.0880 |

**Table 6. Results of DUC2005 for various supervised learning algorithms with Aggregate-Rank-Measure with Neighborhood Rough Sets**

4. **Conclusion**

This paper presents an extension of work in paper [15] and the results are compared based on results from [15] only. Neighbourhood Rough Sets based techniques has been popular in various Machine Learning applications, its use with LERS work especially unsupervised Multi Document Text Summarization was not experimented and theocratized in literature. The present work aims to fill this gap. In this work we analyse the applicability of Rough Set based supervised techniques for Text Summarization. The steps carried out are the following. Firstly, we enhance the popular Rough Set based supervised technique, viz. LERS, using Neighborhood Rough Sets. We compute Neighbourhood Rough based LERS algorithm efficiency for MDTS. Then we use the Aggregate-Rank-Measure [15] for determining the importance of each sentence to the extract.

We have performed experiments with DUC2003 Multi-document dataset and DUC2005 Multi-document datasets while using DUC2002 Multi-document dataset for training. We conclude that Neighborhood Rough Set based techniques are performing better than normal Rough Set based LERS algorithm. Neighborhood Rough Set based proposed technique viz. NLER is improving results considerably when used in combination of Aggregate Rank Measure[15].